\documentclass[runningheads]{llncs}

\usepackage[T1]{fontenc}
\usepackage{graphicx}
\usepackage{amsmath}
\usepackage{amssymb}
\usepackage{amsfonts}
\usepackage{bm}
\usepackage{booktabs}
\usepackage{url}
\usepackage[table]{xcolor}
\usepackage{subcaption}
\usepackage{marvosym}

\graphicspath{{./figs/}}
\begin{document}

\title{Decision-Focused Learning for Mean--Variance Portfolio Optimization via KKT-Based Reformulation}

\titlerunning{Decision-focused Learning for Portfolio Optimization}

\author{Kensei Nosaka\inst{1}\textsuperscript{\Letter}\orcidID{0009-0007-3229-9392} \and
Shunnosuke Ikeda\inst{1}\orcidID{0009-0004-4283-0819} \and
Yuichi Takano\inst{1}\orcidID{0000-0002-8919-1282}}

\authorrunning{K. Nosaka et al.}

\institute{University of Tsukuba, Tsukuba-shi, Ibaraki 305-8573, Japan\\
\email{s2620473@u.tsukuba.ac.jp, ikeda@cs.tsukuba.ac.jp, ytakano@sk.tsukuba.ac.jp}}

\maketitle

\begin{abstract}

Mean--variance portfolio optimization (MVO) is a central framework in data-driven asset management. 
A widely adopted approach is a two-stage framework that first predicts expected returns and then solves the optimization problem based on these predictions, with the predictive models trained by minimizing prediction errors. 
However, this objective of prediction is not aligned with the quality of the downstream portfolio decision. 
Decision-focused learning (DFL), which directly minimizes the downstream decision loss within the learning process, has thus emerged as a promising direction.
However, existing DFL approaches to MVO rely on surrogate losses or constraint relaxations for tractability, creating a structural mismatch between predictive model training and the constrained MVO solved at evaluation.
We propose a single-level optimization formulation that incorporates the Karush--Kuhn--Tucker (KKT) optimality conditions of the lower-level MVO into the upper-level learning problem.
This formulation explicitly preserves the budget and short-sale constraints while remaining tractable for standard nonlinear optimization solvers.
Rolling-window experiments on real-world ETF (Exchange Traded Funds) data across two asset universes with different correlation structures show that our method achieved the best performance on multiple investment metrics and also demonstrated performance improvement due to the proposed regularization.
\end{abstract}

\section{Introduction}

\subsection{Background}

Portfolio optimization is a central problem in asset management.  
Markowitz mean--variance optimization (MVO)~\cite{markowitz1952} is a fundamental framework that formulates the trade-off between expected return and risk.  
In data-driven portfolio management, a two-stage framework is widely used in both practice and research, where expected returns are predicted from historical data and then used as inputs to MVO~\cite{BertsimasKallus2020,lahoud2025}.
However, prediction accuracy is not always aligned with downstream portfolio performance~\cite{butlerkwon2023,mandi2024}.  
In MVO, constraints such as budget and short-sale constraints determine
portfolio allocation structures, and even small prediction errors may lead
to substantial changes in portfolio weights and investment performance
through optimization~\cite{bestgrauer1991,Lee2025}.
This discrepancy partly arises from the inherent separation between prediction and optimization in two-stage frameworks~\cite{BertsimasKallus2020}.

To address this issue, decision-focused learning (DFL) directly minimizes the downstream decision loss rather than the prediction error itself by embedding the optimization problem into the learning process, thereby aligning predictive model training with the downstream decision quality~\cite{mandi2024,Wilder2019}.
Therefore, DFL is expected to improve portfolio performance in MVO.

\subsection{Related Work}

Several studies have investigated DFL for MVO and the influence of decision structures on predictive learning~\cite{Lee2025,mandi2024}.
Elmachtoub and Grigas~\cite{elmachtoubgrigas2022} proposed a framework called Smart Predict-then-Optimize (SPO+), which uses a surrogate loss derived from downstream optimization objectives to emphasize prediction errors that affect decision quality.
However, because SPO+ minimizes a surrogate loss, it does not directly minimize the decision loss. 
Butler and Kwon~\cite{butlerkwon2023} proposed Integrating Prediction and Optimization (IPO), which formulates the joint problem of return prediction and MVO as a bilevel optimization problem.
However, directly minimizing the decision loss leads to a nonconvex optimization problem requiring iterative gradient descent.
Iterative optimization can be avoided by using a closed-form solution to the relaxed MVO problem without short-sale constraints, introducing a structural mismatch between the relaxed MVO for portfolio construction and the constrained MVO for model evaluation.

These existing DFL approaches to MVO therefore rely on surrogate losses or constraint relaxations to improve tractability.  
While computationally efficient, these approximations may fail to fully capture the problem structures governing portfolio allocation in MVO~\cite{Wilder2019}.
Conversely, directly minimizing exact decision losses requires iterative optimization~\cite{amos2017optnet}, which complicates the learning process and increases computational cost.

As a more general approach to DFL, regret minimization has also been studied~\cite{TanFrazier2022}.
Bucarey et al.~\cite{bucarey2024} formulated the exact expected regret minimization as a pessimistic bilevel optimization problem. They further reformulated this bilevel problem into a single-level nonlinear optimization problem using the optimality conditions of the lower-level problem, making it tractable for standard nonlinear optimization solvers.
However, this single-level reformulation technique has not been adapted to the specific constraint structure of MVO, leaving a gap for an MVO-based DFL formulation that reduces to a single-level optimization problem solvable directly by standard nonlinear optimization solvers.

\subsection{Our Contribution}
We propose a single-level nonlinear optimization formulation of MVO-based DFL, derived from the Karush--Kuhn--Tucker (KKT) optimality conditions.
This formulation explicitly preserves the budget and short-sale constraints of MVO during learning.
Specifically, building on the optimality-condition-based reformulation of Bucarey et al.~\cite{bucarey2024}, we formulate the MVO-based DFL as a bilevel optimization problem and convert it to a single-level form using the KKT optimality conditions of the lower-level MVO problem.
In addition, we introduce a regularization scheme that anchors the predictive model parameters to a reference solution, mitigating numerical instability and preventing overfitting.

To evaluate the effectiveness of our method, we conducted computational experiments using real-world ETF (Exchange Traded Funds) data with two asset universes exhibiting different correlation structures.  
Experimental results demonstrate that our method outperformed existing methods on multiple investment metrics.
Additionally, we show that the proposed regularization scheme consistently improved performance.

\section{Proposed Method}
In this section, we first provide a brief overview of the MVO framework. 
We then formulate MVO-based DFL as a bilevel optimization problem and, following Bucarey et al.~\cite{bucarey2024}, reformulate it into a single-level nonlinear optimization problem using the KKT conditions of the lower-level MVO. 
Finally, we introduce a regularization scheme to enhance numerical stability.
We use $[n] := \{1, 2, \dots, n\}$ to denote the set of consecutive positive integers up to $n$.

\subsection{Mean--Variance Portfolio Optimization}

We adopt the MVO framework proposed by Markowitz~\cite{markowitz1952} as the basic portfolio optimization model.
Let $n$ denote the number of assets, $\bm{w}\in\mathbb{R}^n$ the portfolio allocation vector, $\bar{\bm{r}}\in\mathbb{R}^n$ the expected return vector, and $\bm{V}\in\mathbb{R}^{n\times n}$ the covariance matrix.
We assume that $\bm{V}$ is estimated from historical returns and satisfies positive definiteness ($\bm{V}\succ\bm{O}$).
This assumption is naturally satisfied by shrinkage covariance estimators, which produce positive definite estimates~\cite{ledoitwolf2003} and remain well-conditioned even in finite samples~\cite{ledoitwolf2004}.
The feasible region defined by the budget and short-sale constraints is
$
\mathcal{S}
:=
\{
\bm{w}\in\mathbb{R}^n
\mid
\bm{1}^\top\bm{w}=1,\;
\bm{w}\geq\bm{0}
\}.
$

The MVO problem is formulated as
\begin{equation}
\min_{\bm{w}\in\mathcal{S}}
\;
c(\bm{w};\bar{\bm{r}},\bm{V})
:=
\frac{\delta}{2}\bm{w}^\top\bm{V}\bm{w}
-(1-\delta)\bar{\bm{r}}^\top\bm{w},
\label{eq:mvo}
\end{equation}
where $\delta \in (0, 1)$ is a risk-aversion parameter controlling the relative importance of the expected return and risk terms.
Since $\bm{V}\succ\bm{O}$, the objective function is strongly convex, and the problem has a unique global optimum.

Expected returns are estimated using asset-wise predictive models.
Let $\bm{x}_{ti}\in\mathbb{R}^{m}$ denote the feature vector for asset $i\in[n]$ at time period $t\in[T]$, and define the augmented feature vector as
\[
\tilde{\bm{x}}_{ti}
:=
(\bm{x}_{ti}^\top,1)^\top
\in\mathbb{R}^{m+1}.
\]

For simplicity, we employ linear predictive models with regression coefficients $\bm{\theta}_i\in\mathbb{R}^{m+1}$, yielding predicted returns
\[
\hat r_{ti}(\bm{\theta}_i)
:=
\bm{\theta}_i^\top\tilde{\bm{x}}_{ti}.
\]
Defining
\[
\bm{\Theta}
:=
(\bm{\theta}_i)_{i\in[n]}
\in\mathbb{R}^{(m+1)\times n},
\]
the predicted and realized return vectors are respectively defined as
\[
\hat{\bm{r}}_t(\bm{\Theta})
:=
(\hat r_{ti}(\bm{\theta}_i))_{i\in[n]}
\in\mathbb{R}^{n},
\qquad
\bm{r}_t
:=
(r_{ti})_{i\in[n]}
\in\mathbb{R}^{n}.
\]
The predictive model parameters $\bm{\Theta}$ are typically learned by minimizing the sum of squared prediction errors.
We refer to this conventional two-stage framework as prediction-focused learning (PFL)~\cite{mandi2024}.

\subsection{Bilevel Optimization Model}
\label{subsec:bilevel}
We introduce a bilevel optimization model for DFL in which the MVO problem with predicted returns is the lower-level problem, while the upper-level problem determines the parameters $\bm{\Theta}$ that minimize the associated MVO decision loss.

Given the predicted return vector $\hat{\bm{r}}_t(\bm{\Theta})$ and the covariance matrix $\bm{V}_t$, an optimal portfolio allocation $\hat{\bm{w}}_t(\bm{\Theta})$ at time period $t \in [T]$ is defined as
\begin{equation}
\hat{\boldsymbol{w}}_t(\boldsymbol\Theta)
\in
\underset{\boldsymbol{w}_t\in\mathcal{S}}{\arg\min}\;
c\!\left(
\boldsymbol{w}_t; \hat{\boldsymbol r}_t(\boldsymbol\Theta),\boldsymbol{V}_t
\right) .
\label{eq:lower_level_new}
\end{equation}
We also define the ground-truth optimal allocation based on the realized return vector $\bm{r}_t$ as
\begin{equation}
\boldsymbol{w}_t^{\mathrm{oracle}}
\in
\underset{\boldsymbol{w}_t\in\mathcal{S}}{\arg\min}\;
c\!\left(\boldsymbol{w}_t;\boldsymbol{r}_t,\boldsymbol{V}_t\right).
\label{eq:oracle_w}
\end{equation}
Although $\bm{w}_t^{\mathrm{oracle}}$ is unavailable in practical investment settings, it serves as a reference solution for defining the learning objective.

In general, the allocation $\hat{\bm{w}}_t(\bm{\Theta})$ obtained from predicted returns differs from the ground-truth optimal allocation $\bm{w}_t^{\mathrm{oracle}}$.  
We define the decision loss as the gap between these objective values under realized returns:
\begin{equation}
\ell_t(\bm{\Theta})
:=
c\!\left(
\hat{\bm{w}}_t(\bm{\Theta});
\bm{r}_t,
\bm{V}_t
\right)
-
c\!\left(
\bm{w}_t^{\mathrm{oracle}};
\bm{r}_t,
\bm{V}_t
\right).
\label{eq:decision_loss}
\end{equation}
Since the second term does not depend on $\bm{\Theta}$, only the first term is used as the learning objective.

The MVO-based DFL problem is then formulated as
\begin{align}
\min_{\boldsymbol\Theta}
\quad
&\frac{1}{T}\sum_{t=1}^T c\!\left(\hat{\boldsymbol w}_t(\boldsymbol\Theta);\boldsymbol r_t,\boldsymbol V_t\right) 
\label{eq:upper_level_objective}
\\
\text{s.~t.}
\quad
&
\hat{\boldsymbol{w}}_t(\boldsymbol\Theta)
\in
\underset{\boldsymbol{w}_t\in\mathcal{S}}{\arg\min}\;
c\!\left(\boldsymbol{w}_t;\hat{\boldsymbol r}_t(\boldsymbol\Theta),\boldsymbol{V}_t\right), 
~ \forall t \in [T] .
\label{eq:upper_level_constraint}
\end{align}
This problem is difficult to handle directly because it contains the $\arg\min$ operator in the lower-level problem.

\subsection{Single-level Reformulation}
\label{subsec:reformulation}
Following Bucarey et al.~\cite{bucarey2024}, we reformulate the bilevel optimization problem~\eqref{eq:upper_level_objective}--\eqref{eq:upper_level_constraint} as a single-level optimization problem by incorporating the optimality conditions of the lower-level problem.

The lower-level problem~\eqref{eq:upper_level_constraint} minimizes a strongly convex objective function under $\delta>0$ and $\bm{V}_t\succ\bm{O}$, and its feasible region $\mathcal{S}$ is convex.
Thus, the optimal solution is unique.
Moreover, since there exists a feasible solution satisfying $\bm{1}^\top\bm{w}_t=1$ and $\bm{w}_t>\bm{0}$, Slater's condition holds, and the KKT conditions are necessary and sufficient for optimality.

Let $\mu_t\in\mathbb{R}$ and $\bm{\lambda}_t\in\mathbb{R}_{\geq0}^{n}$ denote the Lagrange multipliers associated with the equality constraint $\bm{1}^\top\bm{w}_t=1$ and the short-sale constraint $\bm{w}_t\geq\bm{0}$, respectively.
Replacing the lower-level problem~\eqref{eq:upper_level_constraint} with the corresponding KKT conditions yields the following single-level reformulation:
\begin{align}
\min_{\bm{\Theta},
\{\bm{w}_t,\mu_t,\bm{\lambda}_t\}_{t=1}^{T}}
\quad
&
\frac{1}{T}
\sum_{t=1}^{T}
\left(
\frac{\delta}{2}
\bm{w}_t^\top\bm{V}_t\bm{w}_t
-
(1-\delta)\bm{r}_t^\top\bm{w}_t
\right)
\label{eq:dfl_kkt_obj}
\\
\text{s.~t.}\hphantom{~~~~~}~
\quad
&
\delta\bm{V}_t\bm{w}_t
-
(1-\delta)\hat{\bm{r}}_t(\bm{\Theta})
-
\mu_t\bm{1}
-
\bm{\lambda}_t
=
\bm{0},
\quad
\forall t\in[T],
\label{eq:dfl_kkt_stationarity}
\\
&
\bm{1}^\top\bm{w}_t
=
1,
\quad
\bm{w}_t\geq\bm{0},
\quad
\forall t\in[T],
\label{eq:dfl_kkt_primal}
\\
&
\bm{\lambda}_t
\geq
\bm{0},
\quad
\forall t\in[T],
\label{eq:dfl_kkt_lambda_nonneg}
\\
&
\bm{\lambda}_t
\odot
\bm{w}_t
=
\bm{0},
\quad
\forall t\in[T],
\label{eq:dfl_kkt_complementarity}
\end{align}
where $\odot$ denotes the Hadamard product.

\subsection{Regularization}
The single-level problem~\eqref{eq:dfl_kkt_obj}--\eqref{eq:dfl_kkt_complementarity} is generally nonconvex because it contains complementarity conditions~\eqref{eq:dfl_kkt_complementarity}, and its solution may depend on initialization and numerical optimization settings.
To mitigate this numerical instability and to prevent overfitting, we augment the upper-level objective~\eqref{eq:dfl_kkt_obj} with a squared $\ell_2$-norm regularization term that anchors $\bm{\Theta}$ to a reference parameter $\bm{\Theta}_{\mathrm{ref}}$, yielding the regularized formulation:
\begin{align}
\min_{\bm{\Theta},
\{\bm{w}_t,\mu_t,\bm{\lambda}_t\}_{t=1}^{T}}
\quad
&
\frac{1}{T}
\sum_{t=1}^{T}
\left(
\frac{\delta}{2}
\bm{w}_t^\top\bm{V}_t\bm{w}_t
-
(1-\delta)\bm{r}_t^\top\bm{w}_t
\right)
+
\eta
\left\|
\mathrm{vec}
(
\bm{\Theta}
-
\bm{\Theta}_{\mathrm{ref}}
)
\right\|_2^2
\label{eq:dfl_kkt_obj_reg}
\\
\text{s.~t.}\hphantom{~~~~~~~~}~
&
\text{Eqs.~}\eqref{eq:dfl_kkt_stationarity}\text{--}\eqref{eq:dfl_kkt_complementarity},
\label{eq:dfl_kkt_constr_reg}
\end{align}
where $\eta\geq0$ is a regularization parameter and $\mathrm{vec}(\cdot)$ denotes matrix vectorization.
The reference parameter $\bm{\Theta}_{\mathrm{ref}}$ is assumed to be obtained from a computationally efficient existing predictive method.

\section{Numerical Experiments}
This section evaluates the effectiveness of our method through numerical experiments using real-world ETF data.
We compare portfolio performance against existing methods across two asset universes with different correlation structures and examine the effect of regularization.

\subsection{Experimental Setup}
\label{subsec:experimental_setup}
We used monthly return data from January 2003 to December 2025, computed from adjusted closing prices obtained from Yahoo!~Finance (\url{https://finance.yahoo.com/}), across two ETF universes with different correlation structures.
For an internationally diversified setting, we used eight developed-market country equity ETFs following DeMiguel et al.~\cite{demiguel2009}: 
the United States, Canada, France, Germany, Italy, Japan, Switzerland, and the United Kingdom.
For a highly correlated setting, we considered a sector universe consisting of nine S\&P 500 sector ETFs, corresponding to the Global Industry Classification Standard (MSCI and S\&P Dow Jones Indices, \url{https://www.msci.com/indexes/index-resources/gics}) that were available throughout the experimental period: Information Technology, Financials, Health Care, Energy, Consumer Discretionary, Consumer Staples, Industrials, Utilities, and Materials.
These two universes contrast within-country sectoral exposure with cross-country diversification, and the former is known to exhibit higher inter-asset correlation~\cite{heston1994}.

Model training and portfolio rebalancing were performed in a rolling-window manner.
Input features consisted of 1-, 3-, and 12-month average returns together with 12-month volatility at each rebalancing date ($m=4$), including an intercept term.
All input features were computed using only information available up to the rebalancing date, and the training target for each rebalancing window strictly excluded the current period to prevent look-ahead bias.
At each rebalancing date, the prediction model parameters $\bm{\Theta}$ were trained using the most recent 48 months of data and then fixed for the subsequent one-month investment period.
The training window was shifted forward monthly and the model was retrained repeatedly.
The risk-aversion parameter was fixed at $\delta=0.5$ for all methods.
The covariance matrix $\bm{V}_t$ was estimated by applying oracle approximating shrinkage~\cite{chen2010oas} to an exponentially weighted moving average covariance estimator with decay factor $\alpha=0.97$~\cite{jpm2006riskmetrics}, computed from the most recent 48 months of returns.
All methods used the same covariance estimation procedure.

The regularization parameter $\eta$ was selected via grid search on the validation period from January 2011 to December 2015.
The value yielding the highest validation Sharpe Ratio was selected from $\{0, 0.01, 0.05, 0.1, 0.5, 1, 5, 10, 50, 100, 500\}$.
Final performance was evaluated over the test period from January 2016 to December 2025.

We evaluated investment performance using the following metrics:
\begin{itemize}
  \item \textbf{Sharpe Ratio (SR)}: annualized return divided by annualized volatility.
  \item \textbf{Final Wealth (FW)}: cumulative wealth at the end of the evaluation period, with initial wealth normalized to one.
  \item \textbf{Cumulative Decision Loss (CDL)}: cumulative sum of the decision loss $\ell_t$ in Eq.~\eqref{eq:decision_loss} for $t \in [T]$; lower values indicate higher decision quality.
  \item \textbf{CVaR$_{95}$}: average loss over the worst 5\% of return observations~\cite{rockafellaruryasev2002}.
  \item \textbf{Turnover (TO)}: average portfolio turnover per rebalancing period, as defined in DeMiguel et al.~\cite{demiguel2009}.
\end{itemize}

We compared the performance of the following methods:
\begin{itemize}
  \item \textbf{DFL-KKT}: our KKT-based single-level DFL formulation~\eqref{eq:dfl_kkt_obj_reg}--\eqref{eq:dfl_kkt_constr_reg}.
  \item \textbf{SPO+}: a DFL method minimizing a convex upper bound of decision loss~\cite{elmachtoubgrigas2022}.
  \item \textbf{IPO-CF}: a closed-form DFL method derived from IPO for relaxed lower-level MVO without short-sale constraints~\cite{butlerkwon2023}.
  \item \textbf{IPO-GRAD}: an integrated prediction-and-optimization method using gradient-based updates through the lower-level MVO problem~\cite{butlerkwon2023}.
  \item \textbf{PFL}: a two-stage prediction-focused method that feeds least-squares return predictions into MVO~\cite{lahoud2025,mandi2024}.
  \item \textbf{1/N}: an equal-weight portfolio allocating uniformly across all assets~\cite{demiguel2009}.
  \item \textbf{S\&P 500}: a passive buy-and-hold strategy investing in the S\&P 500 ETF.
\end{itemize}

The DFL-KKT problem~\eqref{eq:dfl_kkt_obj_reg}--\eqref{eq:dfl_kkt_constr_reg} was solved directly using the nonlinear optimization solver KNITRO~15.0 and the optimization modeling language Pyomo in Python.
Each solve used feasibility and optimality tolerances of $10^{-6}$, an iteration budget of up to 20{,}000 iterations, and a per-solve time limit of 500 seconds, with no convergence failures across any reported rebalancing window.
Residuals for the KKT system~\eqref{eq:dfl_kkt_stationarity}--\eqref{eq:dfl_kkt_complementarity} remained below $6.6\times10^{-9}$ across all training configurations, with no violation of dual feasibility.
PFL used asset-wise ordinary least squares regression, and IPO-CF used the closed-form solution of the relaxed MVO.
IPO-GRAD and SPO+ trained their predictive model parameters with Adam at a learning rate of $10^{-3}$ for up to 500 epochs with early stopping (patience of 50).
At each iteration, IPO-GRAD solved the lower-level MVO problem with qpth, a PyTorch-based differentiable quadratic optimization layer, while SPO+ solved the linear program constructed from its surrogate loss with Clarabel~0.11.1.
Since SPO+ was originally designed for linear objective functions, we adopted a reformulation in which the variance term of the MVO objective~\eqref{eq:mvo} is moved to a constraint, with the upper bound $(\bm{w}_t^{\mathrm{oracle}})^\top \bm{V}_t \bm{w}_t^{\mathrm{oracle}}$.
At each rebalancing date, the MVO problem with each method's predictions and the common covariance estimator was solved with Clarabel~0.11.1.

\subsection{Portfolio Performance across Asset Universes}
\label{subsec:portfolio_performance}
We evaluated investment performance through rolling-window backtesting on real market data.
For DFL-KKT, IPO-GRAD, and SPO+, the IPO-CF solution at each rebalancing date served as the warm-start initialization.
For DFL-KKT, the regularization reference parameter $\bm{\Theta}_{\mathrm{ref}}$ was set to the IPO-CF solution at each period. 
The regularization parameter $\eta$ was selected separately for each asset universe according to Section~\ref{subsec:experimental_setup}; the selected values were $\eta=0.5$ for the international universe and $\eta=0.01$ for the sector universe.

Tables~\ref{tab:international_summary} and~\ref{tab:sp500_sectors_summary} give the main evaluation metrics for the international universe and sector universe, respectively.
In each column, the best value across all methods is highlighted in bold.
\begin{table}[t]
\centering
\small
\setlength{\tabcolsep}{6pt}
\renewcommand{\arraystretch}{1.25}
\caption{Performance comparison on the international universe. In each column, the best value is highlighted in bold.}
\label{tab:international_summary}
\begin{tabular}{lrrrrr}
\toprule
Method & SR ($\uparrow$) & FW ($\uparrow$) & CDL ($\downarrow$) & CVaR$_{95}$ ($\downarrow$) & TO ($\downarrow$) \\
\midrule
\rowcolor{gray!10}
DFL-KKT        & \textbf{1.101} & \textbf{4.590} & \textbf{1.644} & 9.180          & 0.991 \\
IPO-GRAD       & 0.984          & 3.869          & 1.719          & \textbf{8.795} & 0.299 \\
IPO-CF         & 0.634          & 2.526          & 1.924          & 11.496         & 1.088 \\
SPO+           & 0.633          & 3.027          & 1.846          & 12.422         & 0.173 \\
PFL            & 0.785          & 3.197          & 1.820          & 10.031         & 1.138 \\
1/N            & 0.747          & 2.774          & 1.902          & 9.481          & 0.016 \\
S\&P 500       & 1.042          & 4.184          & ---            & 9.246          & \textbf{0.000} \\
\bottomrule
\end{tabular}
\end{table}
\begin{table}[t]
\centering
\small
\setlength{\tabcolsep}{6pt}
\renewcommand{\arraystretch}{1.25}
\caption{Performance comparison on the sector universe. In each column, the best value is highlighted in bold.}
\label{tab:sp500_sectors_summary}
\begin{tabular}{lrrrrr}
\toprule
Method & SR ($\uparrow$) & FW ($\uparrow$) & CDL ($\downarrow$) & CVaR$_{95}$ ($\downarrow$) & TO ($\downarrow$) \\
\midrule
\rowcolor{gray!10}
DFL-KKT        & \textbf{1.136} & \textbf{6.490} & \textbf{2.969} & 9.454          & 0.855 \\
IPO-GRAD       & 0.934          & 4.851          & 3.116          & 11.568         & 0.685 \\
IPO-CF         & 0.426          & 1.900          & 3.560          & 14.439         & 1.136 \\
SPO+           & 0.809          & 4.083          & 3.241          & 10.868         & 0.319 \\
PFL            & 0.630          & 2.810          & 3.392          & 11.812         & 1.220 \\
1/N            & 0.931          & 3.480          & 3.321          & 9.450          & 0.025 \\
S\&P 500       & 1.042          & 4.184          & ---            & \textbf{9.246} & \textbf{0.000} \\
\bottomrule
\end{tabular}
\end{table}
On the international universe, DFL-KKT achieved the highest SR and FW and the lowest CDL, with the second-lowest CVaR$_{95}$ after IPO-GRAD.
On the sector universe, DFL-KKT again achieved the highest SR and FW together with the lowest CDL. For CVaR$_{95}$, DFL-KKT (9.454) was substantially lower than the other learning-based methods (i.e., IPO-GRAD, SPO+, PFL) and close to the passive strategies S\&P 500 (9.246) and 1/N (9.450).
The uniformly larger CDL of all methods on the sector universe reflected the higher difficulty of this setting.
In both universes, DFL-KKT exhibited higher turnover than IPO-GRAD and SPO+, though lower turnover than IPO-CF and PFL.

\subsection{Effect of Regularization}
\label{subsec:regularization}

To examine the effect of the regularization term in the DFL-KKT problem~\eqref{eq:dfl_kkt_obj_reg}--\eqref{eq:dfl_kkt_constr_reg}, we varied the regularization parameter $\eta$ and the reference parameter $\bm{\Theta}_{\mathrm{ref}}$, which also served as the warm-start initialization and was set to either the PFL or IPO-CF solution.
Following the grid search procedure in Section~\ref{subsec:experimental_setup}, $\eta=0.5$ was selected for the international universe under both references, while $\eta=0.01$ and $\eta=0.1$ were selected for the sector universe under the IPO-CF and PFL references, respectively.
For the unregularized variant, $\eta$ was set to zero.
The other settings followed Section~\ref{subsec:experimental_setup}.

\begin{figure}[t]
\centering
\includegraphics[width=0.6\textwidth]{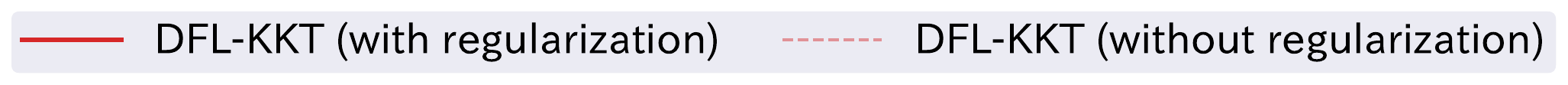}
\vspace{0.1cm}

\begin{subfigure}[b]{0.48\textwidth}
    \centering
    \includegraphics[width=\textwidth,keepaspectratio]{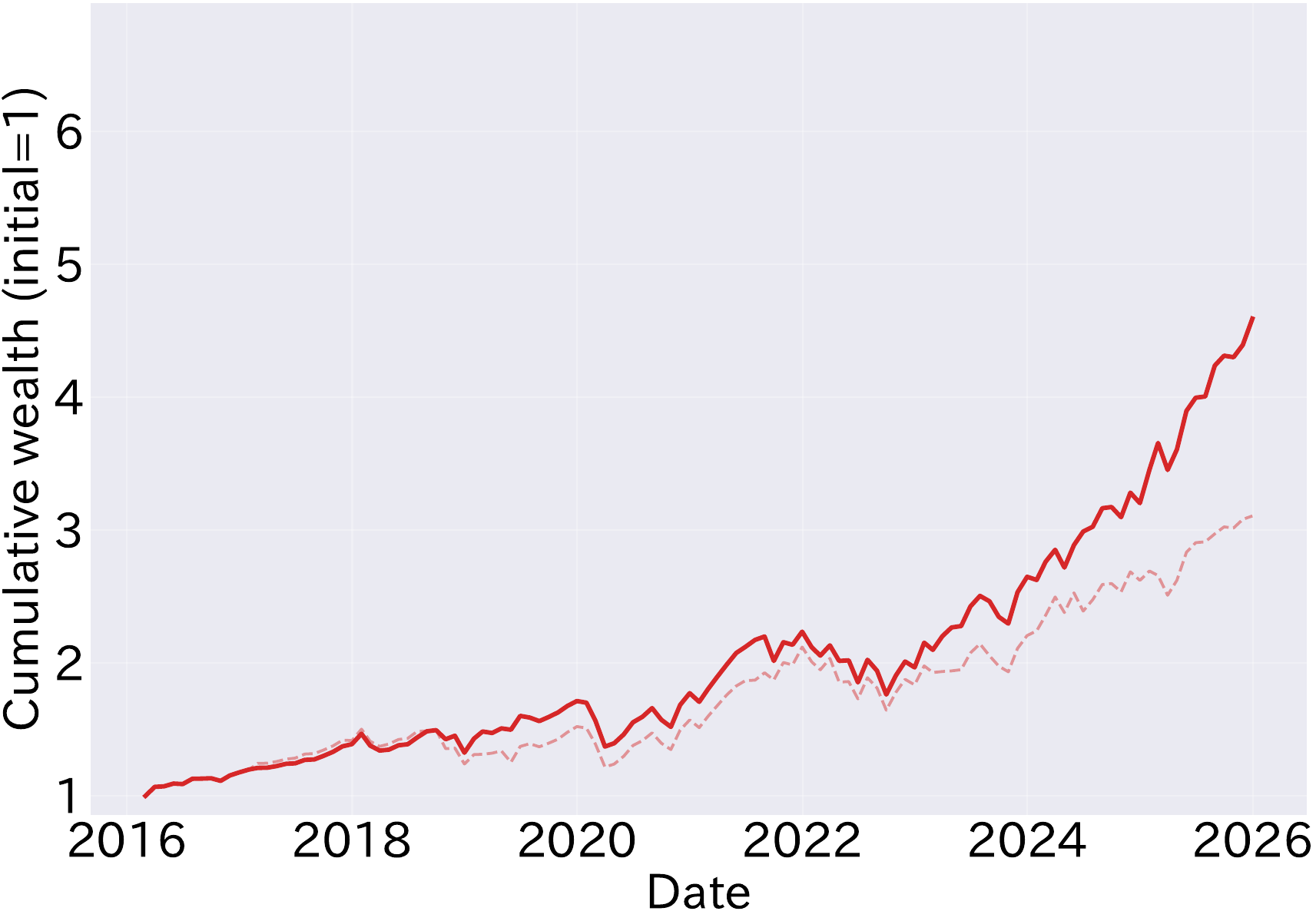}
    \caption{International $\times$ $\bm{\Theta}_{\mathrm{IPO\text{-}CF}}$ ($\eta=0.5$)}
    \label{fig:reg_intl_ipocf}
\end{subfigure}
\hfill
\begin{subfigure}[b]{0.48\textwidth}
    \centering
    \includegraphics[width=\textwidth,keepaspectratio]{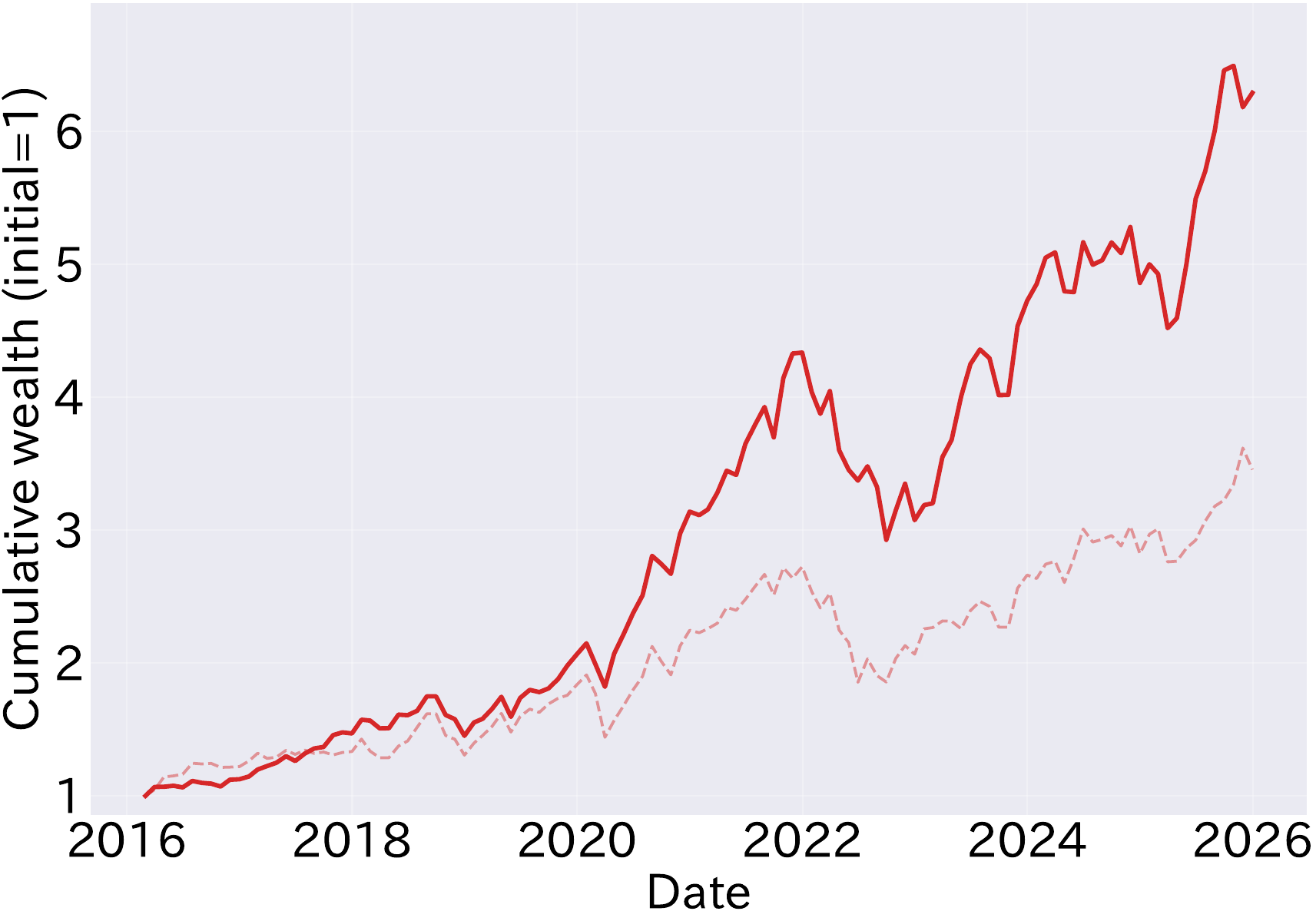}
    \caption{Sector $\times$ $\bm{\Theta}_{\mathrm{IPO\text{-}CF}}$ ($\eta=0.01$)}
    \label{fig:reg_sector_ipocf}
\end{subfigure}
\medskip

\begin{subfigure}[b]{0.48\textwidth}
    \centering
    \includegraphics[width=\textwidth,keepaspectratio]{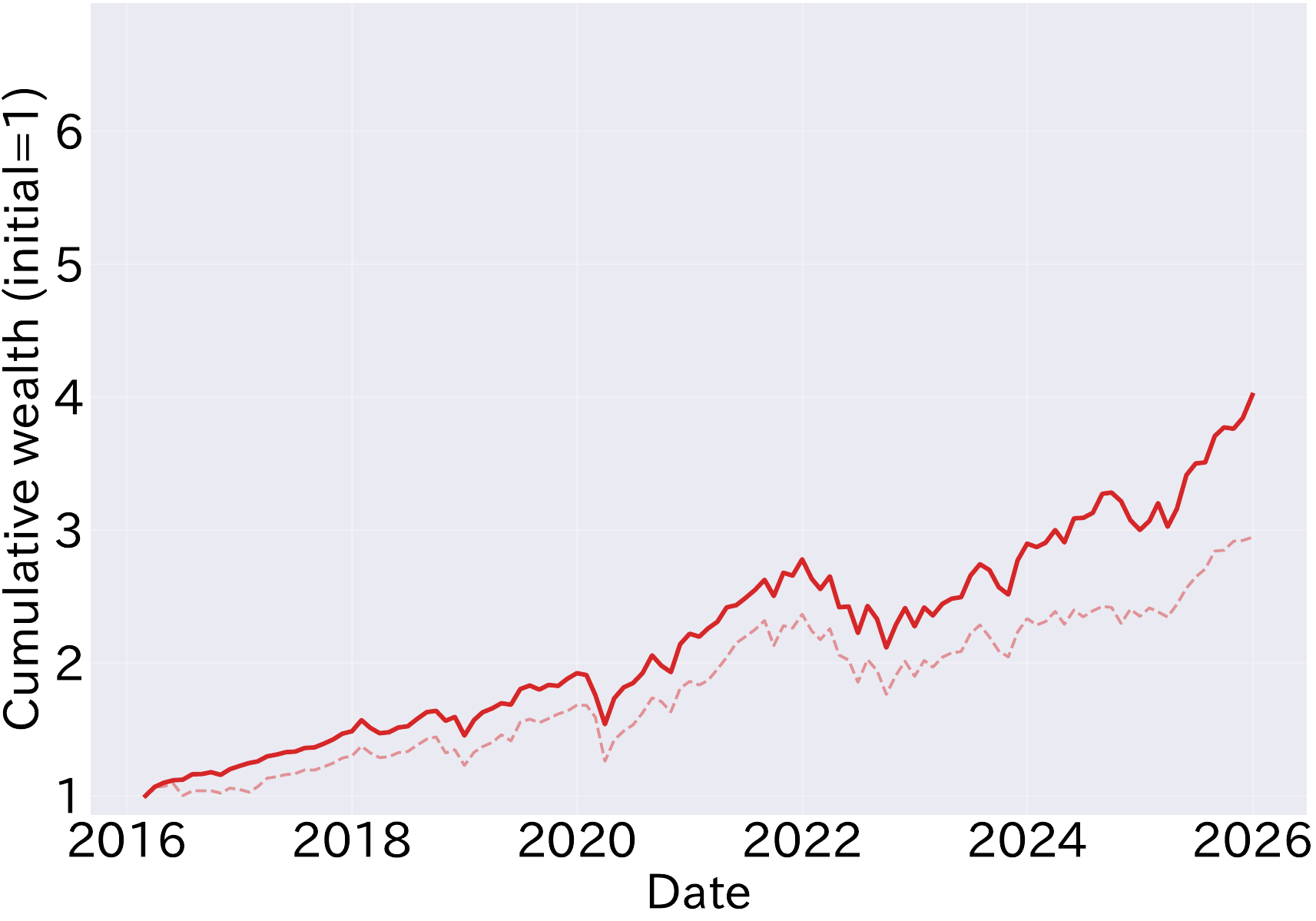}
    \caption{International $\times$ $\bm{\Theta}_{\mathrm{PFL}}$ ($\eta=0.5$)}
    \label{fig:reg_intl_pfl}
\end{subfigure}
\hfill
\begin{subfigure}[b]{0.48\textwidth}
    \centering
    \includegraphics[width=\textwidth,keepaspectratio]{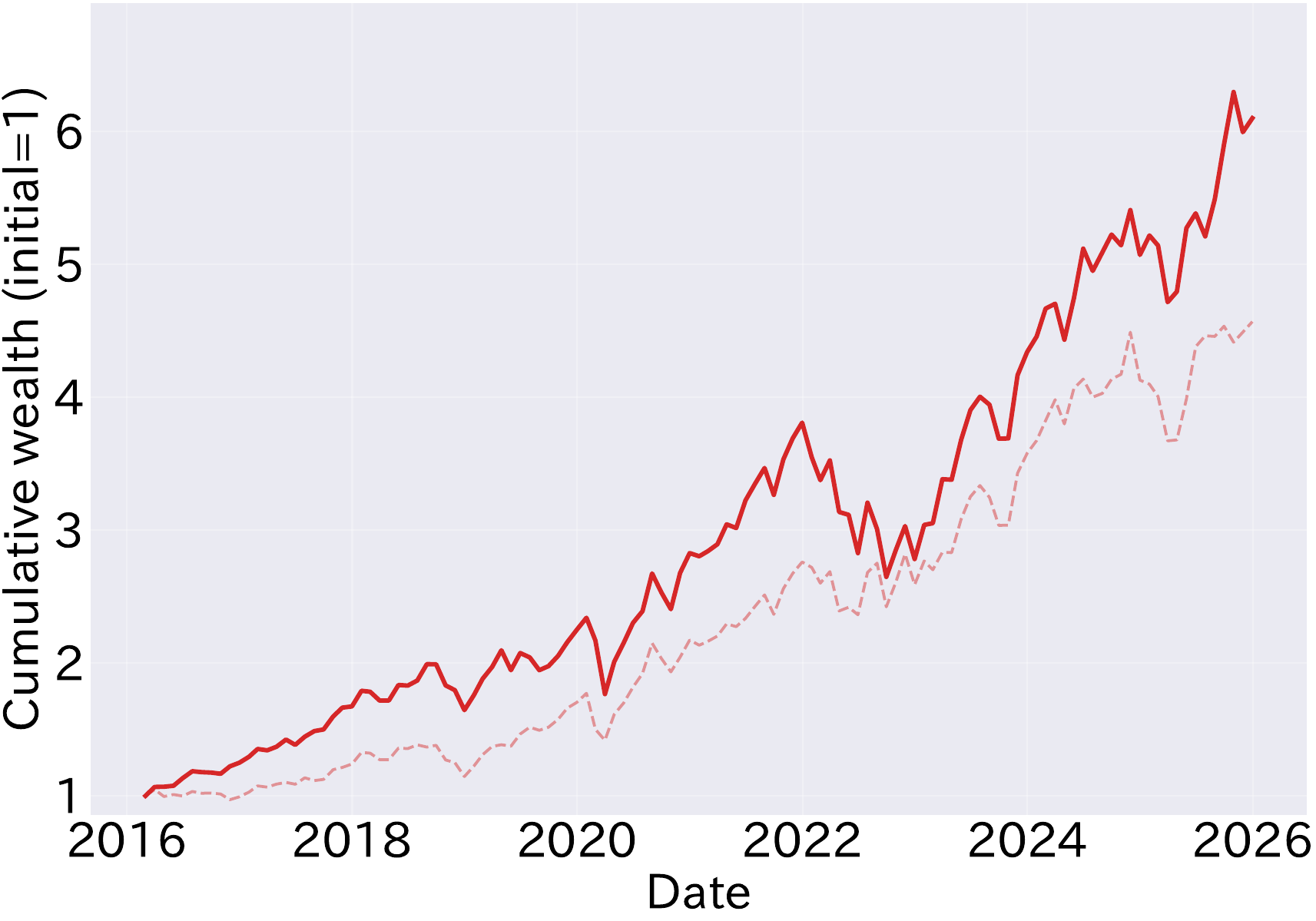}
    \caption{Sector $\times$ $\bm{\Theta}_{\mathrm{PFL}}$ ($\eta=0.1$)}
    \label{fig:reg_sector_pfl}
\end{subfigure}
\caption{Cumulative wealth trajectories of DFL-KKT with and without regularization, under different choices of the reference parameter $\bm{\Theta}_{\mathrm{ref}}$.}
\label{fig:regularization}
\end{figure}

Figure~\ref{fig:regularization} shows that DFL-KKT with regularization consistently outperformed the unregularized variant in both universes, maintaining a higher cumulative wealth for most of the evaluation period under both the PFL and IPO-CF references.
These results suggest that the regularization term stabilized the learning of DFL-KKT by anchoring the learned parameters to the reference, with this benefit robust to the choice of reference solution.

\section{Conclusion}
\label{sec:conclusion}
We proposed a KKT-based single-level nonlinear optimization formulation of DFL for MVO that explicitly preserves the budget and short-sale constraints during learning.
The single-level formulation was derived by replacing the lower-level MVO with its KKT optimality conditions, making it tractable for standard nonlinear optimization solvers.
We further introduced a regularization scheme that anchors the predictive model parameters to a reference solution.

Rolling-window experiments on real-world ETF data across two asset universes with different correlation structures showed that our method achieved the best performance on multiple investment metrics.
The proposed regularization improved performance under different choices of the reference parameter.

Several promising directions remain for future work.
First, developing heuristic or approximation methods would help reduce the computational cost at larger asset dimensions, where the specialized solver may become impractical.
Second, robustness against poor local minima arising from the nonconvex complementarity conditions could be enhanced through advanced initialization or smoothing techniques.
Third, it would be worthwhile to extend the proposed method to incorporate cardinality constraints, coherent risk measures, and robust optimization~\cite{kobayashi2023,takanogotoh2023}.


\bibliographystyle{splncs04}
\bibliography{references}

\end{document}